\pdfoutput=1

\documentclass[letterpaper, 10 pt, conference]{ieeeconf}  

\IEEEoverridecommandlockouts                              
\usepackage{graphics} 
\usepackage{epsfig} 
\usepackage{mathptmx} 
\usepackage{times} 
\usepackage{amsmath} 
\usepackage{amssymb}  
\usepackage{color}
\usepackage{todonotes}
\usepackage{algorithm}
\usepackage{algpseudocode}%
\usepackage{caption}
\usepackage{afterpage}
\usepackage{placeins}
\usepackage{booktabs}
\usepackage{siunitx}
\usepackage{float}
\usepackage{flushend}
\usepackage{algorithmicx}
\usepackage{algorithm}
\usepackage{algpseudocode}
\usepackage{hyperref}
\usepackage[labelfont=bf]{caption}

\title{\LARGE \bf
Exploiting Points and Lines in Regression Forests for RGB-D Camera Relocalization
}

\author{Lili Meng$^{1}$, Frederick Tung$^{1}$, James J. Little$^{1}$, Julien Valentin$^{2}$, Clarence W. de Silva$^{1}$ 
\thanks{$^{1}$ Lili Meng, Frederick Tung, James J. Little and Clarence W. de Silva are with Institute for Computing, Information and Cognitive Systems (ICICS), The University of British Columbia, Vancouver, Canada.}
\thanks{$^{2}$ Julien Valentin is with Perceptive$^{IO}$ Inc, United States.  }
}

\begin{document}

\maketitle
\thispagestyle{empty}
\pagestyle{empty}

\begin{abstract}
Camera relocalization plays a vital role in many robotics and computer vision applications, such as self-driving cars and virtual reality. Recent random forests based methods exploit randomly sampled pixel comparison features to predict 3D world locations for 2D image locations to guide the camera pose optimization. However, these point features are only sampled randomly in images, without considering geometric information such as lines, leading to large errors with the existence of poorly textured areas or in motion blur. Line segments are more robust in these environments. In this work, we propose to jointly exploit points and lines within the framework of uncertainty driven regression forests. The proposed approach is thoroughly evaluated on three publicly available datasets against several strong state-of-the-art baselines in terms of several different error metrics. Experimental results prove the efficacy of our method, showing superior or on-par state-of-the-art performance. 

\end{abstract}

\section{Introduction}
\label{sec:line_introduction}
Camera relocalization plays a vital role in many computer vision, robotics, augmented reality (VR) and virtual reality (AR) applications. In the real world, camera relocalization has empowered the recent consumer robotics products such as Dyson 360 Eye and iRobot Roomba 980 to know where they have previously visited \cite{lukierski2017room}. In AR/VR products such as Hololens and Oculus Rift, camera relocalization helps to correctly overlay visual objects in an image sequence or real world.

Scene Coordinate Regression Forests (SCRF) \cite{shotton2013scene} is the pioneer in using machine learning for camera relocalization. In this method, a regression forest is trained to infer an estimate of each pixel's correspondence to 3D points in the world coordinate. Then these correspondences are used to infer the camera pose with a robust optimization scheme. Since then, various machine learning based methods, mainly random forests based \cite{guzman2014multi, valentin2015exploiting, valentin2016learning, Brachmann_2016_CVPR, LiliRandom, Lili_IROS2017, cavallari2017fly} and deep learning based methods \cite{kendall2015posenet, kendall2016modelling, kendall2017geometric,walch2017image, schmidt2017self,bradski2008learning} have been proposed to accelerate the progress of camera relocalization, in parallel with the traditional but still active feature-based methods \cite{se2005vision, sattler2016efficient} and key-frame based methods \cite{klein2007parallel, glocker2015real}. 

\begin{figure}
\centering
\begin{minipage}{0.44\linewidth}
\centering
\includegraphics[width=\linewidth]{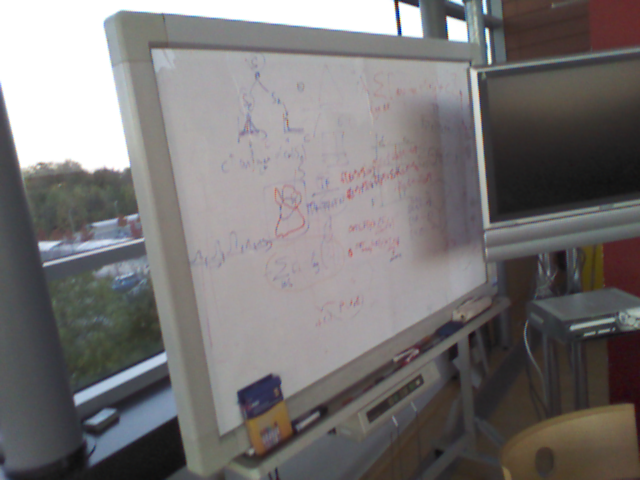}\\
\vspace{-1mm}
(a)
\end{minipage}
\begin{minipage}{0.44\linewidth}
\centering
\includegraphics[width=\linewidth]{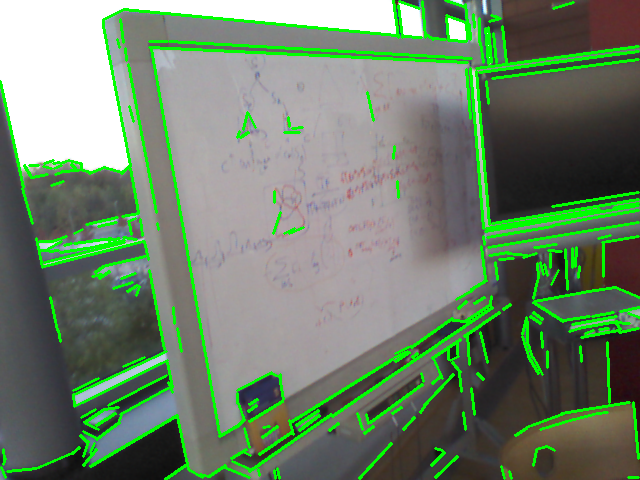} \\
\vspace{-1mm}
(b)  
\end{minipage}  
\vspace{+1mm}
\caption{Line segment example. (a) Input image (b) Line segments. In scenes with little texture and repetitive patterns which are typical in indoor environments, line segments are more robust.}
\label{fig:line_feature_display}
\end{figure}

In these random forests based methods, either RGB-D/RGB pixel comparison features \cite{shotton2013scene, guzman2014multi, valentin2016learning, LiliRandom}, or the sparse features such as SIFT \cite{Lili_IROS2017} are employed, without considering the spatial structure. In poorly textured areas or with the existence of motion blur, line segments provide important geometric information and are robust image features \cite{salaun2016robust} as shown in Fig. \ref{fig:line_feature_display}. Therefore, we propose to use both general points and line-segment points in regression forests to improve camera relocalization performance.  The general points are randomly sampled from the image. The line-segment points are explicitly sampled from line segments in the image. The latter implicitly encodes the line structure of the scene. The main contributions of this work are:

\begin{itemize} 
\item We exploit general points and line-segments points in scene coordinate regression forests, increasing the prediction accuracy.
\item We use the uncertainty of general points predictions and line-segment points predictions to optimize the camera pose.
\item We thoroughly evaluate our methods on three publicly available datasets against several strong state-of-the-art baselines, proving the efficacy of our method with superior or on-par accuracy. 
\end{itemize}

\section{Related work}
Camera relocalization has been widely studied in large scale global localization \cite{nister2006scalable, torii201524, kendall2015posenet}, recovery from tracking failure \cite{klein2007parallel,glocker2015real}, loop closure detection in visual SLAM \cite{whelan2015real, whelan2015elasticfusion} and global localization in mobile robotics \cite{se2005vision, cummins2011appearance},  and sports camera calibration \cite{gupta2011using, chen2018two}. 
Various methods have been proposed to advance camera relocalization performance. We provide a review of random forests based methods here and refer to \cite{Lili_IROS2017} for a review of other methods.

\subsection{Random forests based method for camera relocalization}
In approaches based on random forests for camera relocalization, random forests are used as regressors so we also refer to them as regression forests. These approaches first employ a regression forest to learn each 2D image pixel's corresponding 3D points in the scene's world coordinates with training RGB-D images and their corresponding ground truth poses. Then camera pose optimization is conducted by an adapted version of preemptive RANSAC \cite{shotton2013scene}. Random forests based methods do not need to compute ad hoc descriptors and search for nearest-neighbors, which are time-consuming steps in local feature based and key-frame based methods. Because the environment generally has repeated objects, such as similar chairs in an office room, the random forests have multi-outputs from an input, resulting in ambiguities. To solve this problem, \cite{guzman2014multi} proposed a hybrid discriminative-generative learning architecture to choose optimal camera poses in SCRF. To improve camera relocation accuracy, \cite{valentin2015exploiting} exploits uncertainty from regression forests by using a Gaussian mixture model in leaf nodes. \cite{Brachmann_2016_CVPR, LiliRandom} extend the random forests based method to use a single RGB image in the test stage by employing the Perspective-n-Point method rather than the Kabsch \cite{kabsch1976solution} algorithm in the camera pose optimization stage. However, RGB-D images are still needed in the training stage. \cite{Lili_IROS2017} integrates local features in the regression forests to enable the use of RGB-only images for both training and testing. However, none of these random forests based methods have been evaluated in dynamic scenes for the indoor environment, where the dynamic objects such as people or pets are common.

\subsection{Line segments} Line segment detection and exploitation have been studied for more than three decades and are still very active \cite{burns1986extracting, kahn1990fast, matas2000robust, von2010lsd, salaun2016robust}. Robust gradient orientations of the line segment rather than robust endpoints or gradient magnitudes play a crucial role in the line segment literature \cite{kahn1990fast, von2010lsd, salaun2016robust}. Besides line segment detection, this work is also related to pose estimation using line and/or point features \cite{rehbinder2003pose, salaun2016robust, gupta2011using, lu2015robust}. Unlike methods using line features for direct matching \cite{lu2015robust}, we integrate the line features in the random forests for supervised learning.

\section{Problem setup and method overview}
\label{sec:line_method}
The following three assumptions of the RGB-D camera and input data are made:
(i) the camera intrinsics are known;
(ii) the RGB and depth frames are synchronized;
(iii) the training set contains both RGB-D frames and their corresponding 6 DOF camera pose.

The problem is formulated as: given only a single acquired RGB-D image $\{\mathtt{I, D}\}$, infer the pose $\mathtt{H}$ of an RGB-D camera relative to a known scene.

To solve the problem, we propose to exploit both points and line segments in uncertainty-driven regression forests. Our method consists of two major components. The first component is two regression forests trained using general points and line-segment points respectively. These two forests predict general points and line-segment points in testing. The second component is a camera pose optimization scheme using both point-to-point constraints and point-on-line constraints. 

\section{Regression forests with general points and line-segment points}
\subsection{Points sampling and scene coordinate labels}
To take advantage of the complementary properties of lines and points, our method differentiates the points that are on line segments and general points.

\begin{figure}
\centering
\begin{minipage}{0.44\linewidth}
\centering
\includegraphics[width=\linewidth]{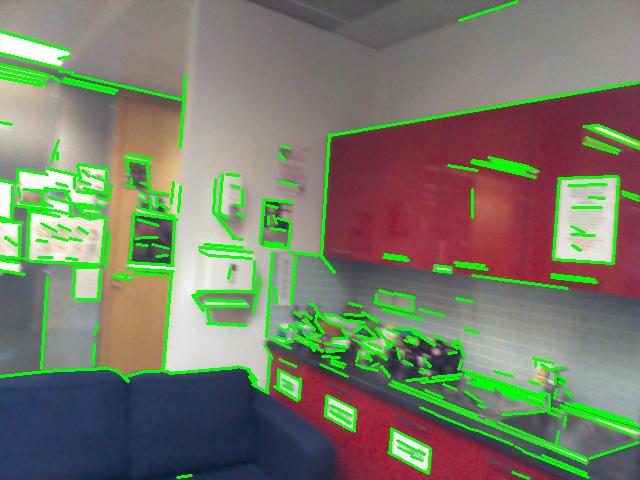}\\
\vspace{-1mm}
(a)
\end{minipage}
\begin{minipage}{0.44\linewidth}
\centering
\includegraphics[width=\linewidth]{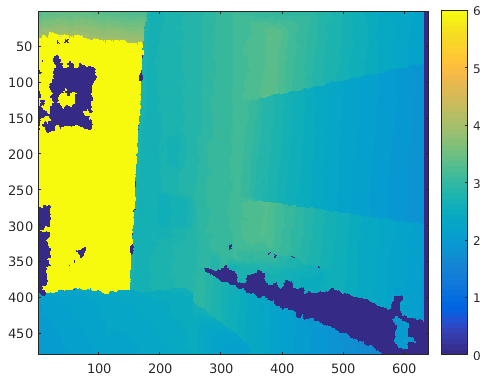} \\
\vspace{-1mm}
(b)  
\end{minipage}  
\vspace{+1mm}
\caption{Depth corruption and discontinuity on line segments. (a) LSD line segments overlaid on original RGB image (b) truncated depth map. Effective depth information is not always available for 2D line segments in the corresponding RGB image, such as the wrong depth values shown on the desk and the glass corridor areas.} 
	\label{fig:corrupted_depth}
\end{figure}

\subsubsection*{\textbf{Line segment sampling}} 
\begin{figure}
	\begin{center}
		\includegraphics[width=0.8\linewidth]{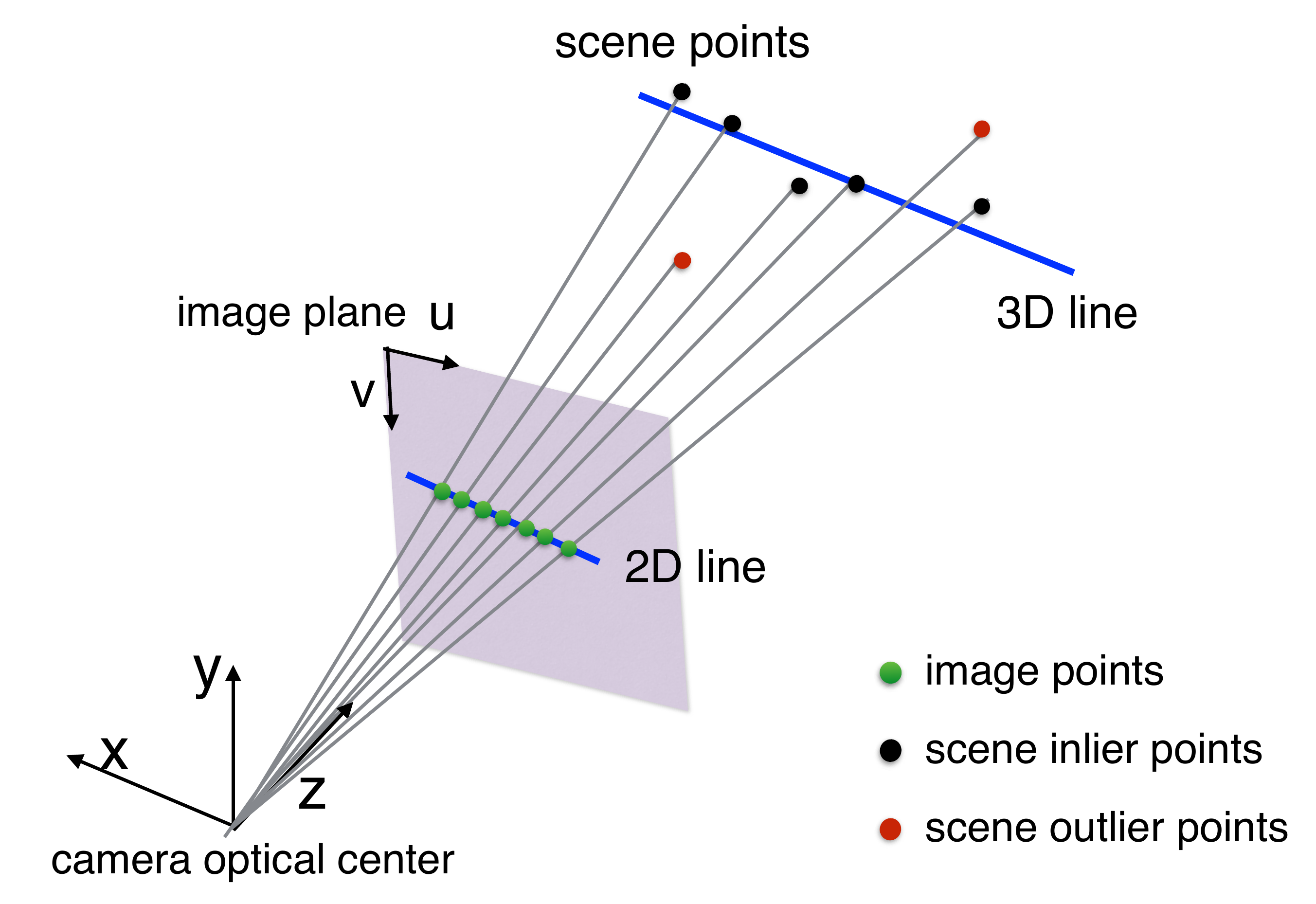}  
	\end{center}
	\vspace{-4mm}
	\caption{3D line estimation based on sampling points. Within a pinhole camera model, the 2D image points are evenly sampled on a 2D image line and then back-projected on the scene coordinate to be 3D scene points. These 3D scene points contain outliers which could be removed by RANSAC by fitting a 3D line in scene coordinates. } 
	\label{fig:3D_line}
\end{figure}

Directly back-projecting the two endpoints to a 3D line using the depth information will cause large errors \cite{lu2015robust} due to discontinuous depth on object boundaries or lack of depth information as shown in Fig. \ref{fig:corrupted_depth}. To avoid this problem, the Line Segment Detector (LSD) \cite{von2010lsd} method is employed to extract a set of 2D line segments $\mathtt{L}=\{\mathtt{l_1}, \mathtt{l_2}, \cdots \}$ from image $\mathtt{I}$ as shown in Fig. \ref{fig:line_feature_display}, and then uniformly sample points from the line as shown in Fig. \ref{fig:3D_line}. Using this method, one could discard the sample points whose depths are unavailable, and only back-project the remaining points to the camera coordinates. The outliers of the back-projected points are further removed by fitting a 3D line using RANSAC \cite{fischler1981random}.

\subsubsection*{\textbf{Random points sampling}} Besides the line-segment points, we also randomly sample points in the image. These two types will be used separately in the training/testing process.

\subsubsection*{\textbf{Image features}} 
We use the pixel comparison feature \cite{shotton2013scene} that associates with each pixel location $\mathbf{p}$:
\begin{equation}
f_{\mathbf{\phi}}(\mathbf{p}) = \mathtt{I} (\mathbf{p}, c_1) - \mathtt{I} (\mathbf{p} + \frac{\mathbf{\delta}}{\mathtt{D}(\mathbf{p})}, c_2)
\label{Eq:Random_RGB_D_feature_PL}
\end{equation}
where $\mathbf{\delta}$ is a 2D offset and $\mathtt{I} (\mathbf{p}, c)$ indicates an RGB pixel lookup in channel $c$. The $\mathbf{\phi}$ contains feature response parameters $\{\mathbf{\delta}, c_1, c_2 \}$. The $\mathtt{D}(\mathbf{p})$ is the depth at pixel $\textbf{p}$ in image $\mathtt{D}$ of the corresponding RGB image $\mathtt{I}$. Pixels with undefined depth and those outside the image boundary are marked and not used as samples for both training and test. In leaf nodes, we also use Walsh-Hadamard Transform (WHT) features \cite{hel2005real} to describe the local patches \cite{Lili_IROS2017}.

\subsubsection*{\textbf{Scene coordinate labels}}
For both general points and line-segment points, the $3D$ points $\mathbf{x}$ in camera coordinate of the corresponding pixel $\mathbf{p}$ are computed by back-projecting the depth image pixels. The scene's world coordinate position $\mathbf{m}$ of the corresponding pixel $\mathbf{p}$ is computed through $\mathbf{m}=\mathtt{H}\mathbf{x}$.

With the present sampling method, one can train both the general point prediction model and line-segment point prediction model in the same way. An alternative method is to use two different models. One model predicts point-to-point correspondences and another model directly predicts line-to-line correspondences. The challenge of the alternative method is that there are few robust and efficient representations of lines in the feature space. Therefore, the proposed method predicts point correspondences and employs point-on-line constraint in the camera optimization process, which greatly simplifies the model learning and prediction process.

\subsection{Regression forest training}
A regression forest is an ensemble of $T$ independently trained decision trees. At this stage, we train a general point regression forest and a line-segment point regression forest. Each tree is a binary tree consisting of split nodes and leaf nodes.

\subsubsection*{\textbf{Weak learner model}}
Each split node $i$ represents a  weak learner parameterized by $\theta_i=\{\mathbf{\phi}_i, \tau_i\}$ where $\mathbf{\phi}_i$ is one feature dimension and $\tau_i$ is a threshold. The tree grows recursively from the root node to the leaf node. At each split node, the parameter $\theta_i$  is sampled from a set of randomly sampled candidates $\Theta_i$. At each split node $i$, for the incoming training set $S_i$,  samples are evaluated on split nodes to learn the split parameter $\theta_i$ that best splits the left child subset $S_i^L$ and the right child subset $S_i^R$ as follows:
\begin{equation}
\label{eq:binary_point_line}
h(\mathbf{p};\theta_i)=
\left\{
\begin{aligned}
0, & \quad \text{if} \ f_{\phi_i}(\mathbf{p}) \leq \tau_i, \quad \text{go to the left subset $S_i^L$.}\\
1, & \quad \text{if} \ f_{\phi_i}(\mathbf{p}) > \tau_i, \quad \text{go to the right subset $S_i^R$.}
\end{aligned}
\right.
\end{equation}
Here, $\tau_i$ is a threshold on feature $f_{\phi_i}(\mathbf{p})$. Although here we use random pixel comparison features as in Eq. \ref{Eq:Random_RGB_D_feature_PL}, the weak learner model can use other general features to adapt to application scenarios such as SIFT feature for outdoor environment as \cite{Lili_IROS2017}.

\subsubsection*{\textbf{Training objective}}
In the training, each split node $i$ uses the randomly generated $\Theta_i$ to greedily optimize the parameters $\theta_i^{*}$ that will be used as the weak learner in the test phase by maximizing the information gain $I_i$:
\begin{equation}
\theta_i^{*} =\arg \max_{\theta_i \in \Theta_i}I_i(\mathtt{S}_i,\theta_i)
\end{equation}
with
\begin{equation}
I_i= E(\mathtt{S}_i) - \sum_{j \in \{L,R\}} \frac{| \mathtt{S}_i^{j}(\theta_i)|}{ |\mathtt{S}_i| } E(\mathtt{S}_i^j(\theta_i))
\end{equation}
where $E(\mathtt{S}_i)$ is the entropy of the labels in $\mathtt{S}_i$, and subset $\mathtt{S}_i^{j}$ is conditioned on the split parameter $\theta_i$. The present work employs a single full-covariance Gaussian model, with the entropy defined as:
\begin{equation}
E(\mathtt{S}) =\frac{1}{2}log({(2\pi e)}^d|\Lambda(\mathtt{S})|) 
\end{equation}
with dimensionality $d=3$ and $\Lambda$ is the full covariance of the labels in $\mathtt{S}$.

At the end of training, all samples reach leaf nodes. In a leaf node, we use the mean shift method \cite{comaniciu2002mean} to estimate a set of modes. Each mode has a mean vector $\mu$ and a covariance matrix $\Lambda$ to described the clustered 3D points. We also store a mean vector of local patch descriptors (i.e.\ WHT features) for each mode. The local patch descriptor will be used to choose the optimal predictions.

\subsection{Regression forest prediction}
In testing, we use the backtracking technique \cite{Lili_IROS2017} to find the optimal prediction within time budgets using a priority queue. In backtracking, the optimal mode has the minimum feature distance from the patch descriptor. To speed up, we use the backtracking number of 8 instead of 16 as in \cite{Lili_IROS2017}.

\section{Camera pose optimization}
\label{sec:camera_pose_uncertainty}
Our method optimizes the camera pose using two types of constraints. The first constraint is point-to-point correspondence. For each sampled camera coordinate point $x_i^c$, the mode is found in $\mathcal{M}_i$ that concurrently best explains the transformed observation $\mathtt{H}x_i^{c}$:
\begin{equation}
(\mu_i^*, \Sigma_i^*) = \arg \max_{(\mu, \Sigma) \in \mathcal{M}_i } \mathcal{N}(\mathtt{H}x_i^c; \mu, \Sigma).
\end{equation}
This can be optimized by minimizing the sum of Mahalanobis distances in world coordinates for image points set $\mathcal{I}_p$:
\begin{equation}
E_{p} = \sum_{i \in \mathcal{I}_p} \|(\mathtt{H}x_i^c - \mu_i^*)^T\Sigma_{x_i}^{*-1}(\mathtt{H}x_i^c - \mu_i^*)\|.
\end{equation}

The second constraint is point-on-line constraint. For each predicted edge point $x_i^w$, the present work transforms its location to the camera coordinate $\mathtt{H}^{-1}x_i^w$ and measures the Mahalanobis distance for line-segment points set $\mathcal{I}_l$ to the associated line $L_i$. This can be optimized by minimizing:
\begin{equation}
E_{l} = \sum_{i \in \mathcal{I}_l} \|(\mathtt{H}^{-1}x_i^w - Q)^T \Sigma_i^{-1}(\mathtt{H}^{-1}x_i^w - Q)\|
\end{equation}
where $Q_i$ is the closest point on $L_i$ to the transformed point $\mathtt{H}^{-1}x_i^w$. The covariance matrix $\Sigma_i$ is rotated from the world coordinate to the camera coordinate by 
\begin{equation}
\Sigma_c = R\Sigma R^T
\end{equation}
where $R$ is the rotation matrix in $\mathtt{H}^{-1}$.

Our method jointly optimizes these two constraints by using the sum over the Mahalanobis distances as the performance index:
\begin{equation}
\mathtt{H}^*  = \arg \min_{H} (E_{p} + E_{l})
\label{equ:joint_calib}
\end{equation}
This energy is optimized by a Levenberg-Marquardt optimizer \cite{more1978levenberg} in a preemptive RANSAC framework as in \cite{shotton2013scene}.  

\section{Experiments}
This section evaluates the developed method on three publicly available datasets for camera relocalization against several strong baselines. 

\subsection{Evaluations on Stanford 4  Scenes dataset}
\label{subsec:Indoor_line_4scenes}

\subsubsection*{\textbf{Dataset}} The $4 \ Scenes$ dataset \cite{valentin2016learning} is with spatial extent of $14$ to $79 m^3$. This large environment includes the offices and apartments which are practical for the application of indoor robot localization. The RGB image sequences were recorded at a resolution of $1296 \times 968$ pixels, and the depth resolution is $640 \times 480$. We re-sampled the RGB images to the depth image resolution to align these images. The ground truth camera poses were from BundleFusion \cite{dai2016bundle}.

\subsubsection*{\textbf{Baselines and error metric}} We use the following state-of-the art methods as baselines: ORB+PnP, SIFT+PnP, Random+SIFT \cite{LiliRandom}, MNG \cite{valentin2016learning}, BTBRF \cite{Lili_IROS2017}, Surface Regression (SR) \cite{brachmann2018learning}. The details on these baselines can be found in \cite{valentin2016learning, Lili_IROS2017,brachmann2018learning}. 

We adopt the commonly used error metric of the proportion of test frames within $5cm$ translational and $5^{\circ} $angular error.
 
\subsubsection*{\textbf{Main results and analysis}}  

\begin{table*}
\begin{center}
\scalebox{1}{
\begin{tabular}{|l|l|cccccc|c|}
\hline
         & \textbf{Spatial}& \multicolumn{6}{c}{\textbf{Baselines}}\vline & \textbf{Ours}  \\ 
  \textbf{Sequence} & \textbf{Extent} & ORB+PnP & SIFT+PnP &Hybrid \cite{LiliRandom}& MNG \cite{valentin2016learning} &BTBRF \cite{Lili_IROS2017} &SR \cite{brachmann2018learning}  &PLForests \\
\hline
\hline
Kitchen &$33m^3$ & 66.39\% & 71.43\% &70.3\% & 85.7\% &92.7\% & \textbf{100}\% &98.9\%\\ 
Living & $30m^3$ & 41.99\% & 56.19\% & 60.0\%&71.6\% & 95.1\% & \textbf{100} \% &\textbf{100}\%\\ 
\hline

Bed     & $14m^3$ & 71.72\% & 72.95\% & 65.7\% &66.4\% &82.8\% & \textbf{99.5}\% &99.0\%\\ 
Kitchen&$21m^3$ & 63.91\% & 71.74\% & 76.7\% &76.7\% & 86.2\%  & \textbf{99.5}\% & 99.0\%  \\  
Living & $42m^3$ & 45.40\% & 56.19\% & 52.2\%&66.6\% & 99.7\% & \textbf{100}\% &\textbf{100}\%\\   
Luke   & $53m^3$ & 54.65\% & 70.99\% & 46.0\%&83.3\%  &84.6\% & 95.5\% &\textbf{98.9}\%\\  
\hline
Floor5a& $38m^3$ & 28.97\% & 38.43\% &49.5\%& 66.2\%  &89.9\% & 83.7\% &\textbf{98.8}\%\\   
Floor5b &$79m^3$ & 56.87\% & 45.78\% & 56.4\%&71.1\% &98.9\%  & 95.0\% &\textbf{99.0}\%\\   
\hline
Gates362 &$29m^3$ & 49.48\% & 67.88\% &67.7\%&51.8\%&96.7\% & \textbf{100}\% &\textbf{100}\%\\    
Gates381& $44m^3$ & 43.87\% & 62.77\% &54.6\%&52.3\% &92.9\% & 96.8\% &\textbf{98.8}\%\\    
Lounge & $38m^3$ & 61.16\% & 58.72\% & 54.0\%&64.2\%&94.8\% & 95.1\% &\textbf{99.1}\%\\    
Manolis  &$50m^3$ & 60.10\% & 72.86\% &65.1\%&76.0\% &98.0\% & 96.4\% &\textbf{100}\%\\    
\hline
\textbf{Average} &--- & 53.7\%& 62.2\% &59.9\%&69.3\% &92.7\% &96.4\% &\textbf{99.3}\%\\   
\hline
\end{tabular}
}
\end{center}
\caption{Camera relocalization results for the 4 Scenes dataset. The percentage of correct frames (within $5cm$ translational and $5^{\circ}$ angular error) of the developed method is shown against six state-of-the-art methods: ORB+PnP, SIFT+PnP, Random+SIFT \cite{LiliRandom}, MNG \cite{valentin2016learning}, SR \cite{brachmann2018learning}. The best performance is highlighted.}
\label{table:rgbd_4scenes_line}
\end{table*}

\begin{figure*}
\centering
\begin{minipage}{0.4\linewidth}
\centering
\includegraphics[width=\linewidth]{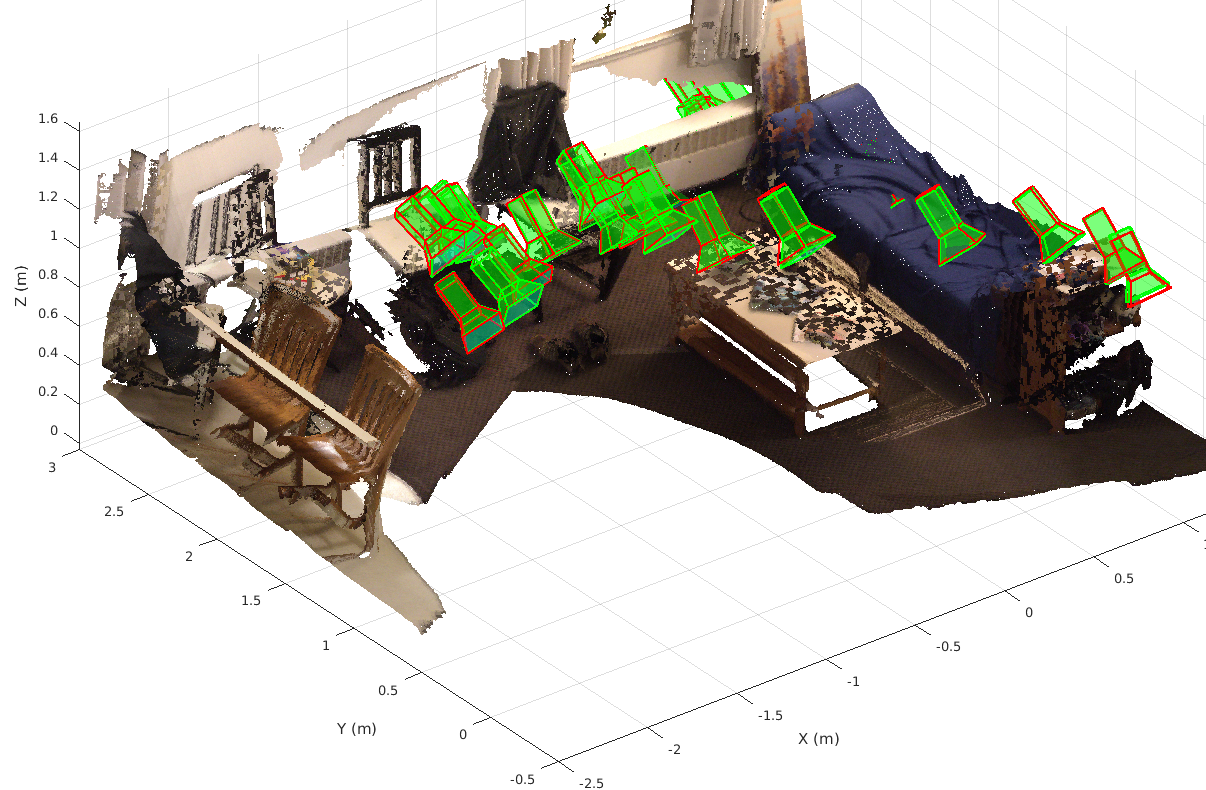}\\
(a)
\vspace{+1mm}
\end{minipage}
\begin{minipage}{0.4\linewidth}
\centering
\includegraphics[width=\linewidth]{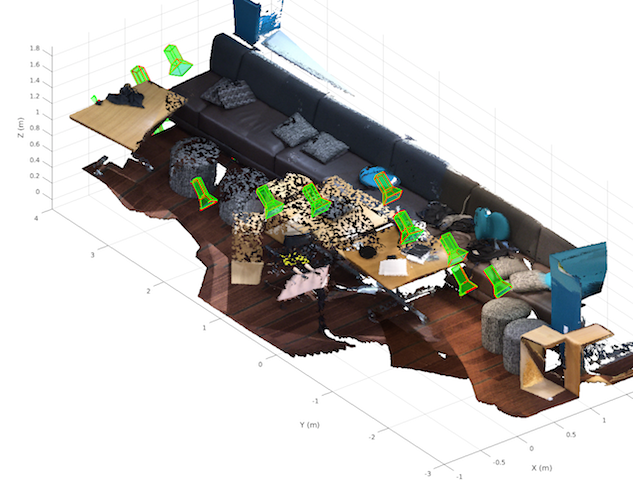} \\
(b)  
\end{minipage}  
\caption{Qualitative results on the 4 Scenes dataset. Best viewed in color. (a) apt1\_living, (b) office2\_5b. We evenly sample every 20 frames in the 3D reconstructed scenes for visualization. The ground truth (hollow red frusta) and the present estimated camera pose (green with light opacity) are very similar. Please note the 3D scenes are only used for visualization purposes and are not used in the present algorithm. }
        \label{fig:4Scenes_good_images_PL}
        \vspace{-2mm}
\end{figure*}

\begin{figure}
\centering
\begin{minipage}{0.41\linewidth}
\centering
\includegraphics[width=\linewidth]{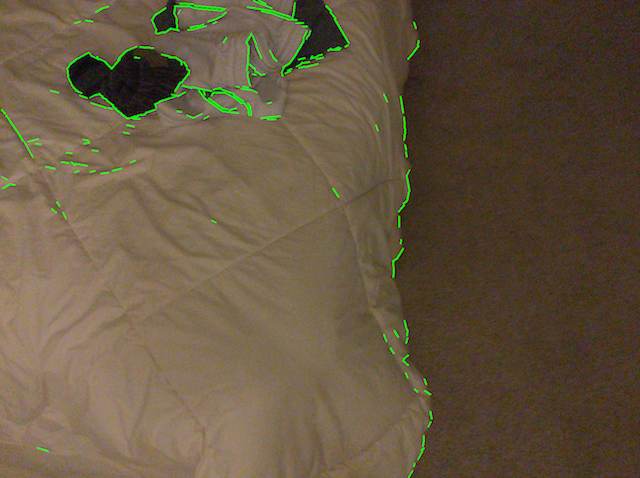}\\
\vspace{+1mm}
(a)
\end{minipage}
\begin{minipage}{0.49\linewidth}
\centering
\includegraphics[width=\linewidth]{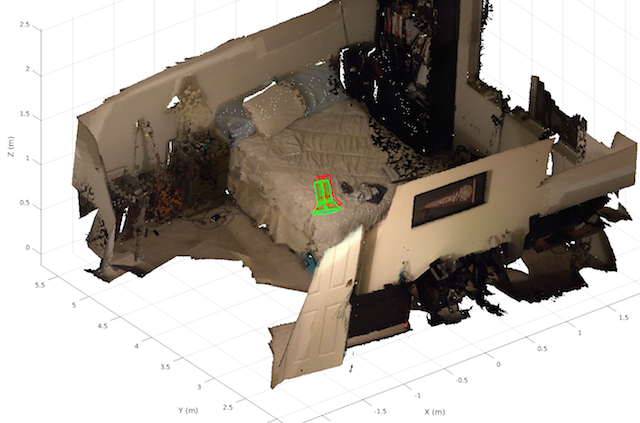} \\
(b)
\end{minipage}  

\caption{Large error examples on Stanford 4 Scenes dataset Apt2\_Luke. (a) an RGB image overlaid with LSD line segments, (b) ground truth (red) and estimated camera pose (green) in the 3D scene. The white sheet and gray floor dominate the scene, little color information and few line segments cause large camera pose error.  }
        \label{fig:4Scenes_failure_images_PL}
        \vspace{-2mm}
\end{figure}

Table \ref{table:rgbd_4scenes_line} shows the main quantitative result on the $4 \ Scenes$ dataset. The proposed method PLForests considerably outperforms all the baselines, with the average correct frame percentage of $99.3\%$. Fig. \ref{fig:4Scenes_good_images_PL} shows some qualitative results of the present camera pose estimation. The estimated camera poses including translations and orientations are very similar to the ground truth camera poses. However, for some scenes, such as Luke, there still exist a few large error camera pose estimates. To further investigate the large error, the RGB images and their large error poses are shown in Fig. \ref{fig:4Scenes_failure_images_PL}. From the RGB images, it is seen that only a very little color information is available from the image. At the same time, the line segments shown in Fig. \ref{fig:4Scenes_failure_images_PL} (a) are not very apparent as well. 

Here we do not compare speed mainly as there is no speed information on the same hardware benchmark. Our current implementation is not optimized for speed and no GPU is used, which makes it possible to speed up with more engineering effort and GPU implementation \cite{sharp2008implementing}. 

\subsection{Evaluations on Microsoft 7 Scenes dataset}
\label{subsec:Indoor_line_7scenes}
The developed method is also evaluated on the Microsoft 7 Scenes dataset.
\subsubsection*{\textbf{Dataset}} The Microsoft 7 Scenes dataset  \cite{shotton2013scene} consists of 7 scenes which were recorded with a handheld Kinect RGB-D camera at $640\times480$ resolution. Each scene includes several camera sequences that contain RGB-D frames together with the corresponding ground-truth camera poses which are obtained from the KinectFusion \cite{newcombe2011kinectfusion} system. The dataset exhibits shape and color ambiguities, specularities and motion blur, which present great challenges for camera relocalization. 

\subsubsection*{\textbf{Baselines and error metric}}  The developed method is compared against SCRF \cite{shotton2013scene}, a multi-output version of SCRF \cite{guzman2014multi}, an uncertainty version of SCRF \cite{valentin2015exploiting} and an autocontext version of SCRF \cite{brachmann2016uncertainty}, Dense correspondence \cite{schmidt2017self} and SR \cite{brachmann2018learning} in terms of correct frame percentage. We also provide results in terms of median translation error and rotation error against Bayesian PoseNet \cite{kendall2016modelling}, PoseNet+Geometric \cite{kendall2017geometric}, CNN+LSTM \cite{walch2017image}, Active Search \cite{sattler2016efficient}, SCRF \cite{shotton2013scene}, BTBRF \cite{Lili_IROS2017}.

\begin{figure}
\centering
\begin{minipage}{1\linewidth}
\centering
\includegraphics[width=\linewidth]{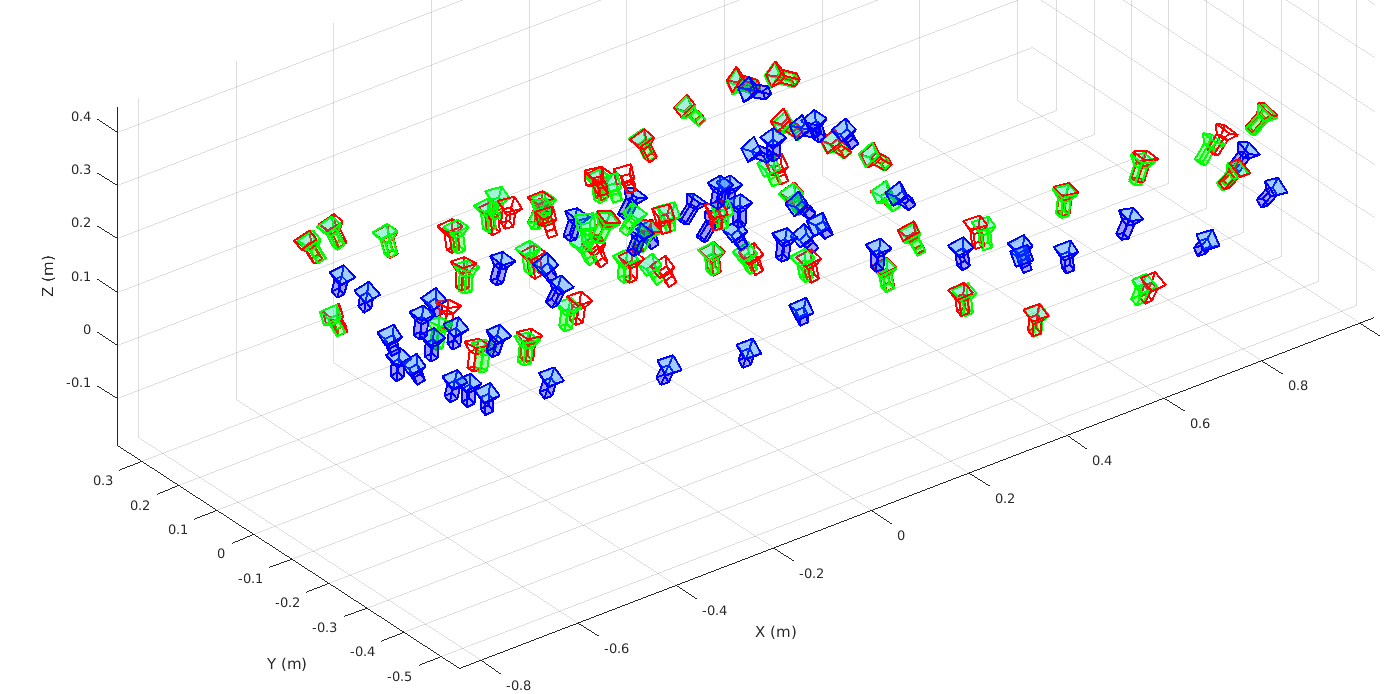}\\
\vspace{+1mm}
\end{minipage}
\caption{Qualitative results for the $Heads$ scene in the 7 Scenes dataset. Best viewed in color and enlarge. Training (blue), test ground truth (red), and test estimated camera poses (green) are evenly sampled for every 20th images. The estimated camera pose is similar to ground truth in both translation and rotation. The large errors occur in places where training poses are very different from test poses.}
        \label{fig:7Scenes_qualitative_images_PL}
        \vspace{-2mm}
\end{figure}

\begin{table*}
\begin{center}
\scalebox{1}{
\begin{tabular}{|l|l|cccccc|c|}
\hline
     &\ &\multicolumn{6}{c}{\textbf{Baselines}}\vline & \textbf{Ours} \\ 
\textbf{Scene} &\textbf{Space}  &SCRF\cite{shotton2013scene} &Multi\cite{guzman2014multi}  &Uncertainty\cite{valentin2015exploiting} &AutoContext \cite{brachmann2016uncertainty} & Dense\cite{schmidt2017self} & SR \cite{brachmann2018learning} &PLForests\\
\hline
\hline 
Chess &$6m^{3}$ &92.6\% & 96\% &99.4\% &\textbf{99.6}\% &97.8\% & 97.6\% &99.5\%\\
Fire &$2.5m^{3}$ &82.9\% &  90\%  &94.6\% &94.0\%& 96.6\%       & 91.9\%  &\textbf{97.6}\%\\
Heads &$1m^{3}$&49.4\%& 56\% &89.3\%&95.9\%&\textbf{99.8}\%     & 93.7\% & 95.5\% \\
Office &$7.5m^{3}$  & 74.9\%& 92\% &\textbf{97.0}\%&93.4\%&\textbf{97.2}\% &87.3\% &96.2\% \\
Pumpkin &$5m^{3}$ &73.7\% & 80\% &\textbf{85.1}\%&77.6\%&81.4\% &61.6\% & 81.4\% \\
Kitchen &$18m^{3}$ &71.8\%& 86\%& 89.3\% &91.1\%&\textbf{93.4}\%& 65.7\% &89.3\% \\
Stairs  &$7.5m^{3}$  &27.8\% & 55\% &63.4\% & 71.7\% &\textbf{77.7}\%& 28.7\% & 72.7\%\\
\hline
Average &--- &67.6\% & 79.3\% &89.5\% &88.1\%& \textbf{92.0}\% & 76.6\% &90.3\%\\
\hline
\end{tabular}
}
\end{center}
\caption{Relocalization results for the 7 Scenes dataset. Test frames satisfying the error metric (within 5cm translational and $5^{\circ}$ angular error) are shown for the proposed method on all scenes against five strong state-of-the-art methods: SCRF \cite{shotton2013scene}, Multi\cite{guzman2014multi},  Uncertainty\cite{valentin2015exploiting}, AutoConext \cite{brachmann2016uncertainty}, Dense\cite{schmidt2017self} and SR \cite{brachmann2018learning}. The best performance is highlighted.}
\label{table:goodPose_7scenes_registered_percentage}
\end{table*}

\begin{table*}
\begin{center}
\scalebox{1}{
\begin{tabular}{|l|cccccc|c|}
\hline
     & \multicolumn{6}{c}{\textbf{Baselines}}\vline &\textbf{Ours} \\ 
\textbf{Scene}  &Geometric \cite{kendall2015posenet} &Bayesian \cite{kendall2016modelling} & CNN+LSTM \cite{walch2017image} &Active Search\cite{sattler2016efficient}& SCRF \cite{shotton2013scene} & BTBRF \cite{Lili_IROS2017} & PLForests\\
\hline
Training & RGB & RGB &RGB &RGB & RGB-D & RGB-D & RGB-D \\ 
Test   & RGB & RGB &RGB &RGB & RGB-D & RGB-D & RGB-D \\
\hline\hline
Chess &0.13m, $4.48^{\circ}$&0.37m, $7.24^{\circ}$ &0.24m, $5.77^{\circ}$ &0.04m, $1.96^{\circ}$& 0.03m, $0.66^{\circ}$&0.015m, $0.59^{\circ}$ &\textbf{0.014}m, $\textbf{0.57}^{\circ}$ \\
Fire  &0.27m, $11.3^{\circ}$&0.43m, $13.7^{\circ}$ &0.34 m, $11.9^{\circ}$ &0.03m, $1.53^{\circ}$ &0.05m, $1.50^{\circ}$&0.016m, $0.89^{\circ}$ &\textbf{0.009}m, $\textbf{0.48}^{\circ}$\\
Heads &0.17m, $13.0^{\circ}$&0.31m, $12.0^{\circ}$ &0.21m, $13.7^{\circ}$ &0.02m, $1.45^{\circ}$& 0.06m, $5.50^{\circ}$&0.020m, $1.84^{\circ}$ &\textbf{0.008}m, $\textbf{0.68}^{\circ}$\\
Office   & 0.19m, $5.55^{\circ}$ &0.48m, $8.04^{\circ}$&
0.30 m, $8.08^{\circ}$ &0.09m, $3.61^{\circ}$ &0.04m, $0.78^{\circ}$& 0.018m, $0.75^{\circ}$ & \textbf{0.017}m, $\textbf{0.73}^{\circ}$ \\
Pumpkin  &0.26m, $4.75^{\circ}$ &0.61m, $7.08^{\circ}$ &0.33 m, $7.00^{\circ}$ &0.08m, $3.10^{\circ}$ &0.04m, $0.68^{\circ}$&0.023m, $0.84^{\circ}$ &\textbf{0.019}m, $\textbf{0.65}^{\circ}$\\
Kitchen  & 0.23m, $5.35^{\circ}$ &0.58m, $7.54^{\circ}$&◦
0.37 m, $8.83^{\circ}$ &0.07m, $3.37^{\circ}$ &0.04m, $0.76^{\circ}$&0.025m, $1.02^{\circ}$  &\textbf{0.025}m, $\textbf{1.02}^{\circ}$ \\
Stairs &0.35m, $12.4^{\circ}$ &0.48m, $13.1^{\circ}$ &0.40 m, $13.7^{\circ}$ &0.03m, $2.22^{\circ}$ & 0.32m, $1.32^{\circ}$&0.040m, $1.22^{\circ}$&\textbf{0.027}m, $\textbf{0.71}^{\circ}$ \\
\hline
Average &0.23m, $8.11^{\circ}$&0.47m, $9.81^{\circ}$&0.31 m, $9.85^{\circ}$ &0.05m, $2.46^{\circ}$&0.08m, $1.60^{\circ}$ & 0.022m, $1.02^{\circ}$ &\textbf{0.017}m, $\textbf{0.70}^{\circ}$\\
\hline
\end{tabular}
}
\end{center}
\caption{Median performance for the 7 Scenes dataset. Results are shown for all scenes against six baselines: PoseNet+Geomeric \cite{kendall2017geometric}, Bayesian PoseNet \cite{kendall2016modelling}, Active Search without prioritization \cite{sattler2016efficient}, SCRF \cite{shotton2013scene}, BTBRF \cite{Lili_IROS2017}. The best performance is highlighted.}
\label{table:goodPose_7scenes_registered_meidan}
\end{table*}

\subsubsection*{\textbf{Main results and analysis}} The main results are shown in comparison with several strong baselines in terms of correct frames in Table \ref{table:goodPose_7scenes_registered_percentage}, and in terms of median performance in Table. \ref{table:goodPose_7scenes_registered_meidan}. The proposed method achieves superior accuracy in terms of median performance, and on-par accuracy in terms of correct frames compared with various methods. Admittedly, Dense \cite{schmidt2017self} achieves a little bit better average accuracy ($1.7\%$), our method still can improve its accuracy with larger backtracking leaf node numbers but at the expense of reducing its speed. Compared with the 4 Scenes dataset, the average accuracy on the 7  Scenes dataset is relatively low. Several reasons may account for this: (i) the training and testing sequences in the 4 Scenes dataset are recorded at the same time as a whole sequence by the same person while the training and test sequences of the 7 Scenes dataset are recorded by different users as different sequences, and split into distinct training and testing sequence sets. (ii) the depth and RGB images have better quality and have better registration in the 4 Scenes. (iii) the scenes in the 7 Scenes are more challenging due to high ambiguity especially in $Stairs$ scene and severe motion blur. Fig. \ref{fig:7Scenes_qualitative_images_PL} shows some qualitative results for the $Heads$ scene of the 7 Scenes dataset. The estimated camera pose is similar to ground truth. Large errors occur in places where the test poses are very different from training poses. Similar findings are also seen in some other scenes.

\subsection{Evaluations on TUM dynamic dataset}
\label{subsec:Indoor_line_TUM}
To demonstrate the performance on datasets that have dynamic objects, the developed method is evaluated using the TUM dynamic dataset. 

\subsubsection*{\textbf{TUM RGB-D dynamic dataset}} TUM RGB-D Dataset \cite{sturm2012benchmark} is mainly for the evaluation of RGB-D SLAM systems. A large set of image sequences of various characteristics (e.g. non-texture, dynamic objects, hand-held SLAM, robot SLAM) from a Microsoft Kinect RGB-D sensor. The ground truth is from a highly accurate and time-synchronized motion capture system. The sequences contain both the color and depth images at an image resolution of $640 \times 480$.
Here, just the dynamic objects dataset is used to complement the previous static 7 Scenes dataset and 4 Scenes dataset where no dynamic objects exist. This dynamic dataset is very challenging as there are severe occlusions and moving objects in the scene. The training set is from the scenes as listed in Table \ref{table:BT_goodPose_TUM_registered_RGBD_line} while the test set is from the respective evaluation sequences. 

\subsubsection*{\textbf{Baselines and error metric}} Three error metrics are used for this evaluation: 'Correct frames' which is the percentage of test frames within $5cm$ and $5^{\circ}$, median translational and angular error, and root mean squared error (RMSE) for Absolute Trajectory Error (ATE) \cite{sturm2012benchmark} which is commonly used in many SLAM system \cite{sturm2012benchmark, mur2015orb}. Unlike the 7 Scenes and 4 Scenes datasets, the training and evaluation data are in different world coordinate systems. The estimated trajectory is still in the training data world coordinate. For alignment, the TUM dataset benchmark provide tools using Horn's method \cite{horn1987closed} to align the estimated trajectory $\mathbf{P}_{1:n}$ and ground truth trajectory $\mathbf{Q}_{1:n}$. The ATE $
\mathbf{F}$ at time step $i$ is computed as 
\begin{equation}
\mathbf{F}_i = \mathbf{Q}_i^{-1}\mathbf{S}\mathbf{P}_i
\end{equation}
The root mean square error (RMSE) over all time indices of the translational components is computed as:
\begin{equation}
RMSE(\mathbf{F}_{1:n})=(\frac{1}{n}\sum_{i=1}^{n}||trans(\mathbf{F}_i)||^2)^{\frac{1}{2}}
\end{equation}
However, this ATE could only be used an auxilliary error metric, as it only considers the translational error while ignoring rotational errors. The translational and rotational error are simultaneously optimized.

\begin{table*}
\begin{center}
\scalebox{1.1}{
\begin{tabular}{|l|l|ccc|ccc|}
\hline 
\textbf{Scene} & \textbf{Trajectory} & \multicolumn{3}{c}{\textbf{Baseline} } \vline &\multicolumn{3}{c}{\textbf{Ours}} \vline \\
\hline
  &    & \multicolumn{3}{c}{BTBRF \cite{Lili_IROS2017} } \vline &\multicolumn{3}{c}{PLForests} \vline \\

& &Correct   &Median & RMSE & Correct &Median & RMSE\\
\hline
\hline 
sitting$\_$static &0.26m &64.6\% &0.015m, $0.99^{\circ}$ & 0.018m &\textbf{72.2}\% & \textbf{0.012}m, $\textbf{0.93}^{\circ}$ &\textbf{0.016}m\\
sitting$\_$xyz  & 5.50m &70.2\%  &0.029m, $0.72^{\circ}$ & \textbf{0.039}m&\textbf{74.1}\% &\textbf{0.028}m, $0.73^{\circ}$&0.047m\\
sitting$\_$halfsphere & 6.50m &\textbf{44.4}\% & \textbf{0.056}m, $\textbf{1.59}^{\circ}$ &\textbf{0.046}m &39.8\% & 0.061m, $1.76^{\circ}$& 0.072m \\
sitting$\_$rpy & 1.11m & 74.6\% &0.029m, $0.98^{\circ}$ & \textbf{0.065}m &\textbf{77.9}\% &\textbf{0.026}m, $\textbf{0.94}^{\circ}$&0.069m \\
walking$\_$halfsphere &7.68m & \textbf{61.7}\% & \textbf{0.042}m, $\textbf{1.03}^{\circ}$ &\textbf{0.085}m &46.6\% & 0.052m, $1.26^{\circ}$ &0.111m \\
walking$\_$rpy &2.70m &\textbf{53.7}\% & 0.047m, $1.14^{\circ}$ &0.551m &53.4\% &0.047m, $\textbf{1.01}^{\circ}$ &\textbf{0.169}m\\
walking$\_$static &0.28m & 89.2\% &0.019m, $0.49^{\circ}$ &0.027m& \textbf{97.3}\% &\textbf{0.017}m, $\textbf{0.35}^{\circ}$ &$\textbf{0.021}$m\\
walking$\_$xyz&5.79m &41.7\% &  0.048m, $1.24^{\circ}$ &0.064m&\textbf{46.6} \% &\textbf{0.047}m, $\textbf{1.22}^{\circ}$ &\textbf{0.063}m\\

\hline
Average & & 62.5\% &0.036m, $1.02^{\circ}$ & 0.119m &\textbf{63.5}\% &0.036m, $1.02^{\circ}$ & \textbf{0.071}m \\
\hline
\end{tabular}
}
\end{center}
\caption{Camera relocalization results for the TUM dynamic dataset. Performances are shown using three different error metrics: correct percentage, median, RMSE of ATE. }
\label{table:BT_goodPose_TUM_registered_RGBD_line}
\end{table*}

\begin{figure}
\centering
\begin{minipage}{0.49\linewidth}
\centering
\includegraphics[width=\linewidth]{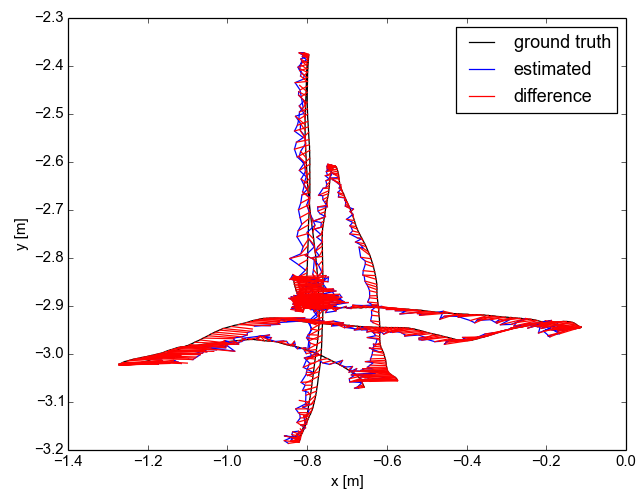}\\

(a)
\end{minipage}
\begin{minipage}{0.49\linewidth}
\centering
\includegraphics[width=\linewidth]{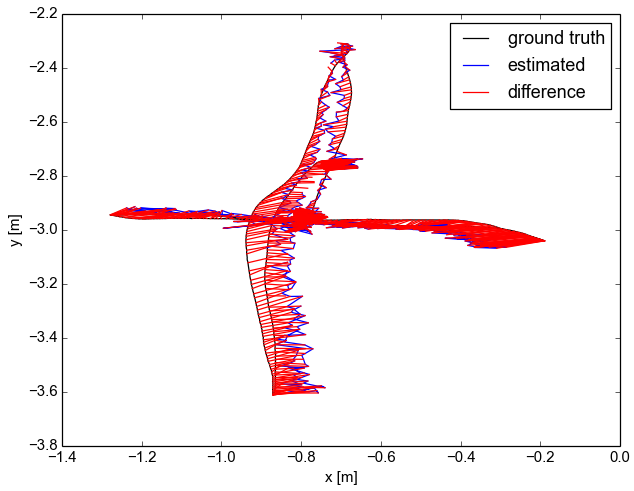} \\

(b)  
\end{minipage}  
 
\begin{minipage}{0.49\linewidth}
\centering
\includegraphics[width=\linewidth]{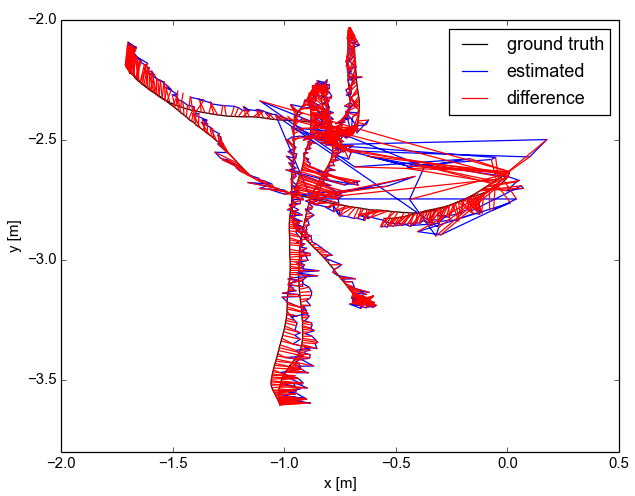} \\
(c) 
\end{minipage} 
\begin{minipage}{0.49\linewidth}
\centering
\includegraphics[width=\linewidth]{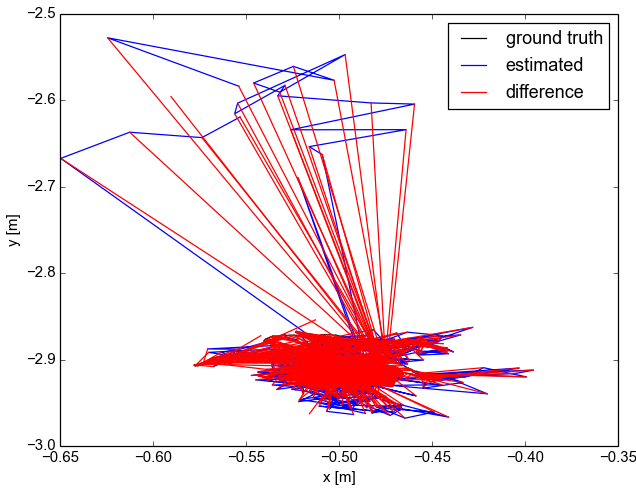}\\
(d)
\end{minipage}

\caption{Quantitative results on $TUM$ dynamic dataset. Two good scenes and two bad scenes are shown. (a) sitting\_xyz, (b) walking\_xyz, (c) walking\_halfsphere, (d) sitting\_rpy. }
        \label{fig:TUM_gt_estimated_PL}
     
\end{figure}

\begin{figure}
\centering
\begin{minipage}{0.49\linewidth}
\centering
\includegraphics[width=\linewidth]{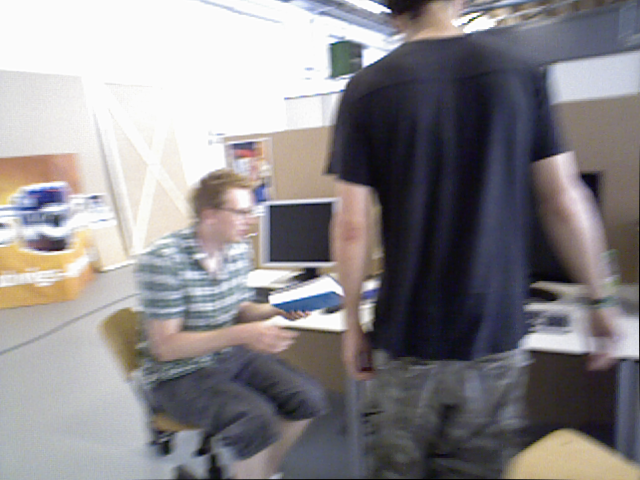}\\
\vspace{-1mm}
(a)
\end{minipage}
\begin{minipage}{0.49\linewidth}
\centering
\includegraphics[width=\linewidth]{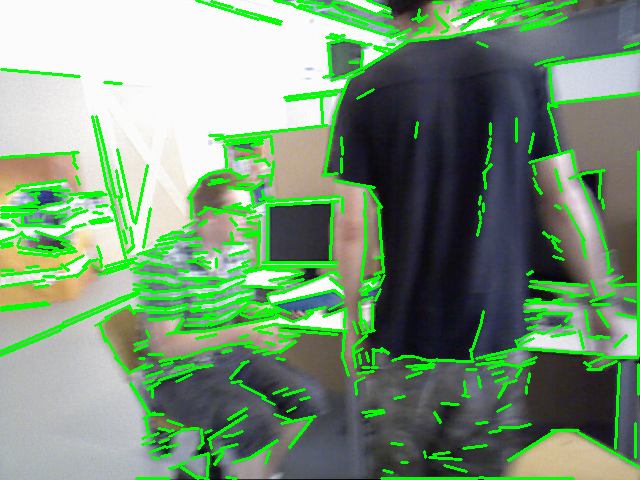} \\
\vspace{-1mm}
(b)  
\end{minipage}  
\begin{minipage}{0.49\linewidth}
\centering
\includegraphics[width=\linewidth]{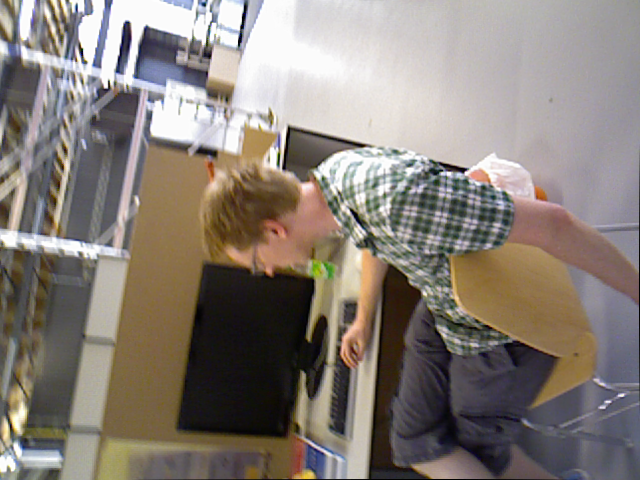}\\
\vspace{-1mm}
(c)
\end{minipage}
\begin{minipage}{0.49\linewidth}
\centering
\includegraphics[width=\linewidth]{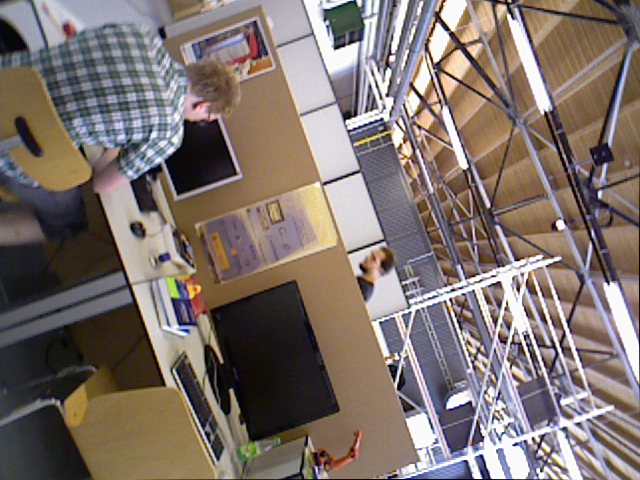} \\
\vspace{-1mm}
(d)  
\end{minipage}  
\caption{Failure cases on TUM dynamic dataset. (a) walking\_halfsphere. Dynamic objects dominates the image and severe motion blur exists. (b) walking\_halfsphere, overlaid with LSD line segments, there are too many line segments on dynamic objects. (c) sitting\_halfsphere, (d) walking\_rpy. Large rotation angle changes.}
        \label{fig:TUM_failure_images_PL}
        \vspace{-2mm}
\end{figure}

\subsubsection*{\textbf{Main results and analysis}} Table \ref{table:BT_goodPose_TUM_registered_RGBD_line} shows the main results of the proposed method on the TUM Dynamic dataset. Compared with the static datasets, our method works with a lower accuracy due to high dynamics. Our method is more accurate than BTBRF in terms of average correct frames and RMSE of ATE, and has the same average median performance. The proposed method could work satisfactorily under highly dynamic scenes and other challenges but still struggles in some extreme cases. Fig. \ref{fig:TUM_gt_estimated_PL} shows the qualitative results on two good and two bad sequences. In these bad sequences, dynamic occlusions dominate the scene, which causes too few inliers and therefore lead to failure cases, such as shown in Fig. \ref{fig:TUM_failure_images_PL} (a). The other important failure case is large rotation angle changes, as shown in Fig. \ref{fig:TUM_failure_images_PL} (c) (d). This is because the random RGB comparison feature is not rotation invariant. Although our PLForests performs much better in static scenes such as 7 Scenes and 4 Scenes than BTBRF, there is no significant difference between BTBRF and PLForests in the TUM dynamic scenes in terms of correct frames and median performance. It shows that a different error metric matters. One possible reason is that there are too many line segments on dynamic objects such as shown in Fig.\ref{fig:TUM_failure_images_PL} (b), but they are not stable line segments in correspondence matching and can not be filtered as outliers by RANSAC in this situation.

\subsection{Implementation details}
The proposed method is implemented with C++ using OpenCV on an Intel 3GHz i7 CPU, 16GB memory Mac system. The parameter settings for our PLForests are: point tree number $T_p=5$ and line-segment point tree number $T_l=5$; 500 training images per tree; 5,000 randomly sampled pixels per training image; the maximum depth of trees is 25; the maximum backtracking leaves is 8.
\section{Conclusions}
In this work, we propose to exploit both line and point features within the framework of uncertainty-driven regression forests. We simultaneously consider the point and line predictions in a unified camera pose optimization framework. We extensively evaluate the proposed approach in three datasets with different spatial scale and dynamics. Experimental results demonstrate the efficacy of the developed method, showing superior state-of-the-art or on-par performance. Furthermore, different failure cases are thoroughly demonstrated, throwing some light into possible future work.  For future work, it will be promising to improve our current method to be more robust to dynamic environments such as using weighted edge points \cite{li2017rgb}. Moreover, implementing our current method on a GPU will be more efficient \cite{sharp2008implementing} and it will also be interesting to integrate our method for a robot autonomous navigation system \cite{autonomous_IROS2017}.




\bibliographystyle{IEEEtran}
\bibliography{ICRA2018}

\end{document}